\documentclass[lettersize,journal]{IEEEtran}
\usepackage{amsmath,amsfonts,amssymb,booktabs}
\usepackage{algorithmic}
\usepackage{algorithm}
\usepackage{array}
\usepackage[caption=false,font=normalsize,labelfont=sf,textfont=sf]{subfig}
\usepackage{textcomp}
\usepackage{stfloats}
\usepackage{url}
\usepackage{verbatim}
\usepackage{graphicx}
\usepackage{cite}
\begin{document}

\title{Minimising Biasing Word Errors for Contextual ASR with the Tree-Constrained Pointer Generator}

\author{Guangzhi Sun,~\IEEEmembership{Student Member,~IEEE,} Chao Zhang,~\IEEEmembership{Member,~IEEE,} Philip C. Woodland,~\IEEEmembership{Fellow,~IEEE}
}



\maketitle

\begin{abstract}
 Contextual knowledge is essential for reducing speech recognition errors on high-valued long-tail words. This paper proposes a novel tree-constrained pointer generator (TCPGen) component that enables end-to-end ASR models to bias towards a list of long-tail words obtained using external contextual information. With only a small overhead in memory use and computation cost, TCPGen can structure thousands of biasing words efficiently into a symbolic prefix-tree, and creates a neural shortcut between the tree and the final ASR output to facilitate the recognition of the biasing words. 
To enhance TCPGen, we further propose a novel minimum biasing word error (MBWE) loss that directly optimises biasing word errors during training, along with a biasing-word-driven language model discounting (BLMD) method during the test. All contextual ASR systems were evaluated on the public Librispeech audiobook corpus and the data from the dialogue state tracking challenges (DSTC) with the biasing lists extracted from the dialogue-system ontology. Consistent word error rate (WER) reductions were achieved with TCPGen, which were particularly significant on the biasing words with around 40\% relative reductions in the recognition error rates. MBWE and BLMD further improved the effectiveness of TCPGen, and achieved more significant WER reductions on the biasing words. TCPGen also achieved zero-shot learning of words not in the audio training set with large WER reductions on the out-of-vocabulary words in the biasing list.
\end{abstract}


\section{Introduction}
\IEEEPARstart{E}{nd-to-end} ASR systems often suffer from high recognition errors on long-tailed words that are rare or not included in the training set. Contextual biasing integrates external contextual knowledge into ASR systems at test-time, which plays an increasingly important role in addressing the long-tail word problem in many applications \cite{shallow_context_1,shallow_context_2,shallow_context_3,deep_context_1,deep_context_2,deep_context_3,deep_context_4,deep_context_5,deepshallow,DBRNNT,ne_correction,unsupervised_context,word_mapping,lm_pointer}. Contextual knowledge is often represented as a list (referred to as a \textit{biasing list}) of words or phrases (referred to as \textit{biasing words}) that are likely to appear in a given context. There exist a variety of resources from which biasing lists can be extracted or organised, such as a user's contact book or playlist, recently visited websites and the ontology of a dialogue system \textit{etc}. Although biasing words occur infrequently and hence may only have a small impact on the overall word error rate (WER), they are mostly content words, such as nouns or proper nouns, and are thus particularly important to downstream tasks and highly valuable. A word is more likely to be recognised if it is incorporated in the biasing list, which makes contextual biasing critical to the correct recognition of those rare content words in an end-to-end ASR system.

As end-to-end ASR systems \cite{e2e_attention_1, e2e_rnnt_1} are often designed to incorporate all of the required knowledge into a single static model, it is particularly challenging for such systems to integrate contextual knowledge specific to a dynamic test-time context. Therefore, dedicated contextual biasing approaches have been proposed, including shallow fusion (SF) with a special weighted finite-state transducer (WFST) or a language model (LM) adapted for the contextual knowledge \cite{shallow_context_1,shallow_context_2,shallow_context_3,lm_pointer,unsupervised_context,word_mapping}, attention-based deep context approaches \cite{deep_context_1,deep_context_2,deep_context_3,deep_context_4,deep_context_5}, as well as deep biasing (DB) with a prefix tree for improved efficiency when dealing with large biasing lists \cite{deepshallow,DBRNNT}. More recently, contextual biasing components with a neural shortcut that directly modifies the output distribution has been proposed \cite{TCPGen, MEM}, which can be jointly optimised with ASR systems.

In this paper, a tree-constrained pointer generator (TCPGen) component is proposed and developed for end-to-end contextual speech recognition. This paper extends our work in \cite{TCPGen}. TCPGen directly interpolates the original model distribution with an extra distribution (the TCPGen distribution) estimated from contextual knowledge,  based on a dynamic interpolation weight predicted by the TCPGen component. TCPGen creates a neural shortcut between biasing lists and the final ASR output distribution to allow end-to-end training of a single neural network model. 
In contrast to the original work on pointer generators \cite{pointer_1,pointer_2,pointer_3} which rely on the attention-mechanism to attend to all biasing words, TCPGen represents biasing lists as wordpiece-level symbolic prefix trees and only attends to the valid subset of the biasing words at each time step in decoding, and can thus handle large biasing lists with high efficiency. Furthermore, TCPGen also uses wordpieces instead of whole words for pointer generators, which allows the entire system to use wordpieces to address the out of vocabulary (OOV) issue. Therefore, as not only a few frequent words but also OOV words exist in our biasing lists, TCPGen can be viewed as a structure that achieves zero-shot learning of previously unseen words without changing model parameters. 
As a result, TCPGen combines the advantages of both neural and symbolic methods and improves pointer generators for ASR applications with large biasing lists. 

To further improve the effectiveness of contextual biasing with TCPGen, a minimum biasing word error (MBWE) loss is proposed to directly optimise the model performance on the biasing words. By changing the risk function of the widely used minimum word error (MWE) loss \cite{mwe0,mwe1,mwe2,mwe3,mwe4}, 
the proposed MBWE loss has a greater focus on minimising the expected errors of the rare and OOV words in the biasing list associated with that utterance. An efficient beam search algorithm is proposed for MBWE training in RNN-T by limiting the number of output wordpieces at each encoder step to one \cite{nsc1,nsc2}. 
Moreover, to address the issue that end-to-end models often suffer from the internal LM effect that biases towards common words, a biasing-driven LM discounting (BLMD) method is proposed. The BLMD method extends the density ratio-based LM discounting method \cite{density_ratio} to incorporate an additional discounting factor for the TCPGen distribution before interpolation.

In this paper, TCPGen, a generic component for end-to-end ASR systems, is integrated into both attention-based encoder-decoder (AED) \cite{e2e_attention_1,e2e_attention_2,e2e_attention_3,e2e_attention_4,e2e_attention_5,LAS} and recurrent neural network transducer (RNN-T) \cite{e2e_rnnt_1,e2e_rnnt_2,e2e_rnnt_3,e2e_rnnt_4,e2e_rnnt_5}. MBWE and BLMD methods are also applied to both types of end-to-end models in conjunction with TCPGen. Experiments were performed on two different types of data, including the Librispeech audiobook data and goal-oriented dialogue data. Improvements in word error rate (WER) were achieved by using TCPGen in both AED and RNN-T configurations across all test sets compared to the baseline and the DB method, with a significant WER reduction on the biasing words. 

The remainder of this paper is organised as follows. Sec. \ref{sec:related} reviews related work. Sec. \ref{sec:tcpgen} introduces the TCPGen component. Sec. \ref{sec:mbr} and \ref{sec:ilmd} describe MBWE training and BLMD for TCPGen respectively. Sec. \ref{sec:exp} and \ref{sec:results} present the experimental setup and results. Sec. \ref{sec:conclusion} gives the conclusions.

\section{Related work}
\label{sec:related}
\subsection{End-to-end contextual speech recognition}
Various contextual biasing algorithms have been recently developed for end-to-end ASR. One of the major streams of research in this area focuses on representing biasing lists as extra WFST which is incorporated into a class-based LM via SF \cite{shallow_context_1,shallow_context_2,shallow_context_3,word_mapping}. Such methods usually rely on special context prefixes like ``call'' or ``play'', which limits its flexibility in handling more diverse grammars in natural speech. Neural network-based deep context approaches using attention mechanisms have also been proposed. These approaches encode a biasing list into a vector to use as a part of the input to the end-to-end ASR models \cite{deep_context_1,deep_context_2,deep_context_3,deep_context_4,deep_context_5}. Although the deep context approaches address the dependency issue on syntactic prefixes in the SF methods, they are more memory intensive and less effective for handling large biasing lists.

Work in \cite{deepshallow} jointly adopted the use of deep context and the SF of a WFST together in an RNN-T. It also improved the efficiency by extracting the biasing vector from only a subset of wordpieces constrained by a prefix tree representing the biasing list, which is referred to as deep biasing (DB) in this paper. Work in \cite{DBRNNT} extended the prefix-tree-based method to RNN LMs which are used for SF to achieve further improvements in biasing words. Moreover, while previous studies only focused on industry datasets, researchers in \cite{DBRNNT} proposed and justified a simulation of the contextual biasing task on open-source data by adding a large number of distractors to the list of biasing words in an utterance. More recently, \cite{TCPGen,MEM} have simultaneously proposed creating a neural shortcut between the biasing list and the final model output distribution. 

\subsection{MWE training for end-to-end ASR models}

The MWE loss that directly minimises the expected WER \cite{mwe0}, has become increasingly popular in training end-to-end ASR models\cite{mwe1,mwe2,mwe3,mwe4,mwe5,mwe6,mwe7}. MWE training using the N-best hypotheses to approximate the expected word errors has been proposed in \cite{mwe1} for the AED model, which was then improved in \cite{mwe4}. More recently, work in \cite{mwe6} applied LM fusion and internal LM estimation during MWE training of an AED model to improve the N-best approximation. Work in \cite{mwe5} exploited a lattice structure in place of the N-best list to calculate the expected word errors. For RNN-T models, \cite{mwe2} applied the same N-best approximation as in AED to calculate the expected errors. Work in \cite{mwe3} further improved MWE training by relating the gradient calculation for each alignment of a hypothesis in the N-best list to the original RNN-T loss function, which enabled offline decoding of N-best lists when abundant CPU resource was available. Moreover, work in \cite{phil2002} first discussed the necessity of LM discounting in sequence discriminative training for HMM-based large-vocabulary continuous speech recognition, and \cite{mwe7} applied MWE to the hybrid auto-regressive transducer (HAT) \cite{HAT}. 

\subsection{LM discounting}

Various LM discounting algorithms have recently been proposed to minimise the internal LM effect of end-to-end ASR models, in particular when applying the model to a test set in a different domain from the training data. In \cite{density_ratio}, a density ratio method was introduced to estimate the score from a source-domain external LM that is to be subtracted from the target-domain LM score. Later, the HAT was proposed to preserve the modularity of a hybrid system and allowed internal LM scores to be estimated and discounted \cite{HAT}. An internal LM estimation method was proposed in \cite{ilme_1} to estimate the source-domain LM score directly from the end-to-end ASR model. This method was further improved by performing internal LM training \cite{ilmt_1} in order to better estimate the internal LM score. 
Moreover, \cite{rnntjoint} proposed regularising the internal LM in RNN-T training to avoid over-fitting to text priors. 

\section{Tree-constrained pointer generator}
\label{sec:tcpgen}

\begin{figure}[h]
    \centering
    \includegraphics[scale=0.35]{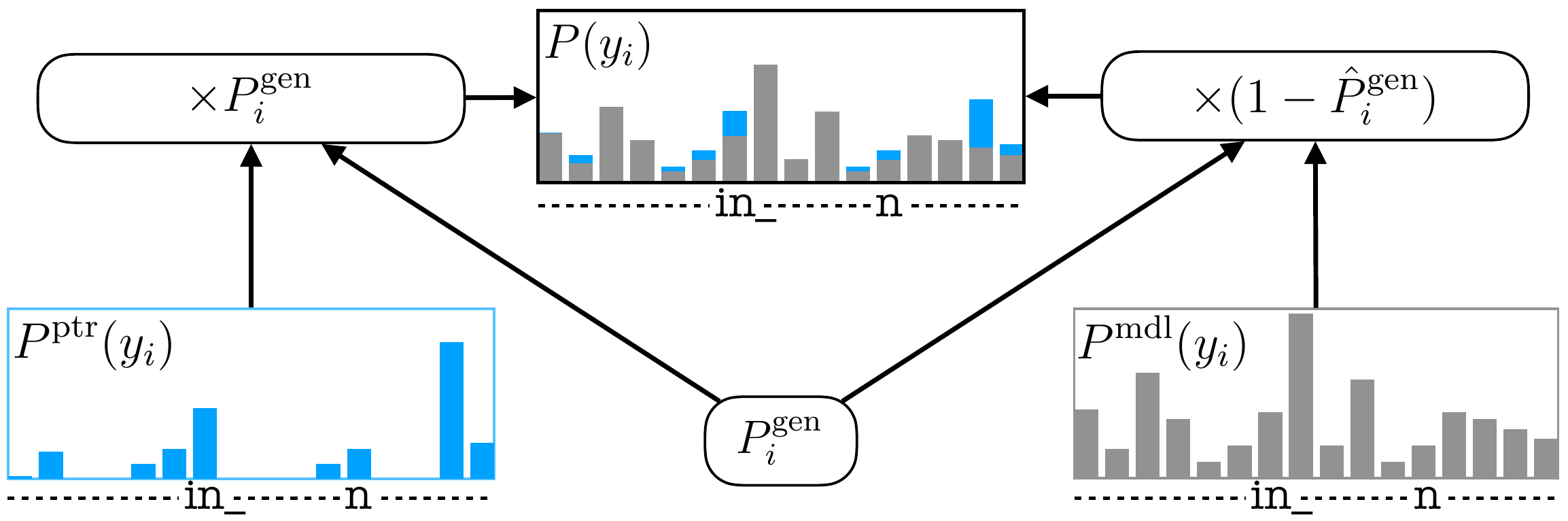}
    \caption{Illustration of interpolation in TCPGen with corresponding terms in Eqn. (\ref{eq:TCPGen_final}). $P^\text{ptr}(y_i)$ is the TCPGen distribution. $P^\text{mdl}(y_i)$ is the distribution from a standard end-to-end model. $P(y_i)$ is the final output distribution. $\hat{P}^\text{gen}_i$ and $P^\text{gen}_i$ are the scaled and unscaled generation probabilities.}
    \label{fig:interpolation}
\end{figure}

TCPGen is a neural network-based component combining symbolic prefix-tree search with a neural pointer generator for contextual biasing, which also enables end-to-end optimisation. TCPGen represents the biasing list as a wordpiece-level prefix tree. At each output step, TCPGen calculates a distribution over all valid wordpieces constrained by the prefix tree. TCPGen also predicts a generation probability which indicates how much contextual biasing is needed at a specific output step. The final output distribution is the weighted sum of the TCPGen distribution and the original AED or RNN-T output distribution (Fig. \ref{fig:interpolation}).

\begin{figure}[h]
    \centering
    \includegraphics[scale=0.35]{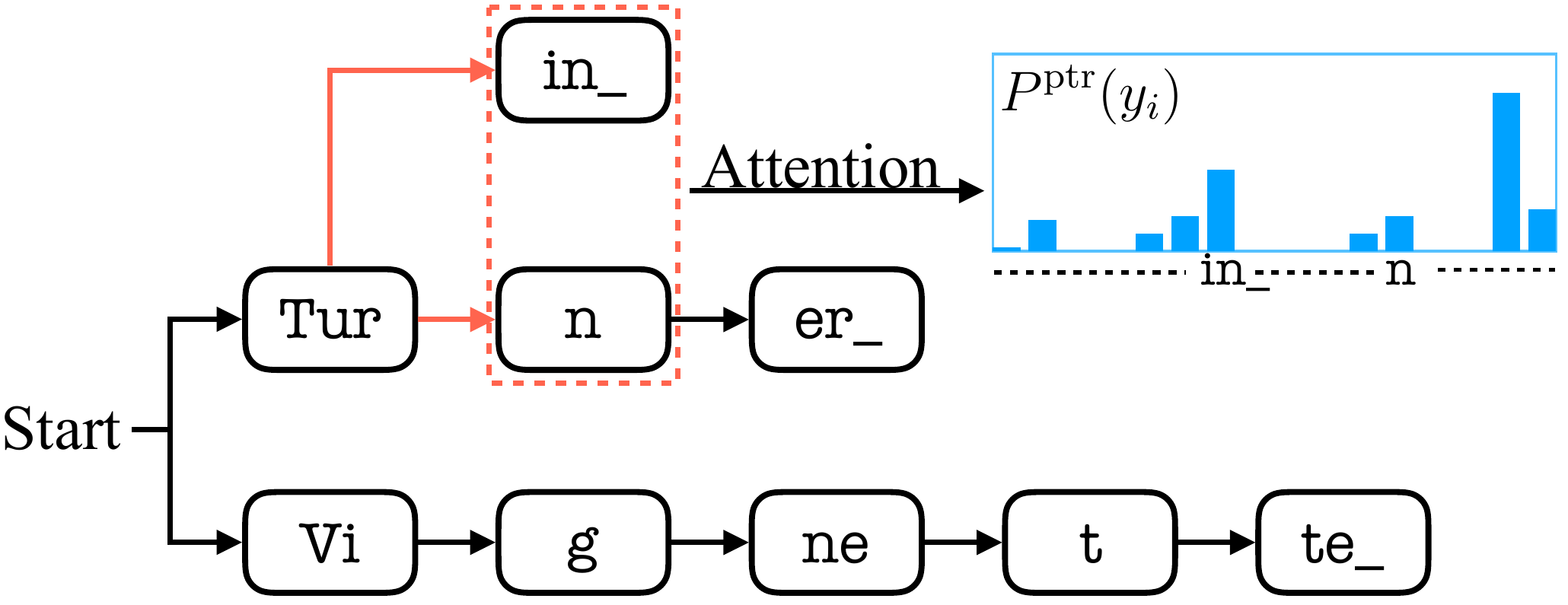}
    \caption{An example of prefix tree search and attention in TCPGen. With previous output \texttt{Tur}, \texttt{in\_} and \texttt{n} are two valid wordpieces on which attention will be performed. A word end unit is denoted by \texttt{\_}.}
    \label{fig:tree}
\end{figure}

The key symbolic representation of the external contextual knowledge in TCPGen is the prefix tree. For simplicity, examples and equations in this section are presented for a specific search path, which can be generalised easily to beam-search with multiple paths. In the example prefix tree with three biasing words shown in Fig. \ref{fig:tree}, if the previously decoded wordpiece is \texttt{Tur}, wordpieces \texttt{in\_} and \texttt{n} form the set of valid wordpieces $\mathcal{Y}^\text{tree}_i$. 
Denoting $\mathbf{x}_{1:T}$ and $y_i$ as input acoustic features and output wordpiece, $\mathbf{q}_i$ as the query vector carrying the decoding history and acoustic information, $\mathbf{K}=[...,\mathbf{k}_j,...]$ as the key vectors, a scaled dot-product attention is performed between $\mathbf{q}_i$ and $\mathbf{K}$ to compute the TCPGen distribution $P^\text{ptr}$ and an output vector $\mathbf{h}^{\text{ptr}}_i$ as shown in Eqns. \eqref{eq:TCPGen_attention} and \eqref{eq:TCPGen_value}.
\vspace{-0.1cm}
\begin{equation}
    P^{\text{ptr}}(y_{i}|y_{1:i-1},\mathbf{x}_{1:T}) = \text{Softmax}(\text{Mask}(\mathbf{q}_i\mathbf{K}^\text{T}/\sqrt{d})),
    \label{eq:TCPGen_attention}
    \vspace{-0.1cm}
\end{equation}
\begin{equation}
    \mathbf{h}^{\text{ptr}}_i = \sum\nolimits_{j} P^{\text{ptr}}(y_i=j|y_{1:i-1},\mathbf{x}_{1:T})\,\mathbf{v}^\text{T}_j,
    \label{eq:TCPGen_value}
    \vspace{-0.1cm}
\end{equation}
where $d$ is the size of $\mathbf{q}_i$ (see \cite{transformer}), Mask$(\cdot)$ sets the probabilities of wordpieces that are not in $\mathcal{Y}^{\text{tree}}_i$ to zero, and $\mathbf{v}_j$ is the value vector relevant to $j$. For more flexibility, an \textit{out-of-list} (OOL) token is included in $\mathcal{Y}^{\text{tree}}_i$ indicating that no suitable wordpiece can be found in the set of valid wordpieces. To ensure that the final distribution sums up to 1, the generator probability is scaled as $\hat{P}^\text{gen}_i=P^\text{gen}_i(1-P^\text{ptr}(\text{OOL}))$, and the final output can be calculated as shown in Eqn. \eqref{eq:TCPGen_final}.
\begin{equation}
    P(y_i) = P^{\text{mdl}}(y_i)(1-\hat{P}^\text{gen}_i) + P^{\text{ptr}}(y_i)P^\text{gen}_i,
    \label{eq:TCPGen_final}
\end{equation}
where conditions, $y_{1:i-1}, \mathbf{x}_{1:T}$, are omitted for clarity. $P^{\text{mdl}}(y_i)$ represents the output distribution from the standard end-to-end model, and $P^{\text{gen}}_i$ is the generation probability. 

\subsection{TCPGen in AED}
A standard AED contains three components: an \textit{encoder}, a \textit{decoder} and an \textit{attention network}. 
The encoder. encodes the input, $\mathbf{x}_{1:T}$, into a sequence of high-level features, $\mathbf{h}^{\text{enc}}_{1:T}$. At each decoding step $i$, 
an attention mechanism is first used to combine the encoder output sequence into a single context vector, $\mathbf{c}_i$. The decoder computation is thus as follows:
\begin{equation}
    \mathbf{h}^{\text{dec}}_i = \text{Decoder}(\mathbf{y}_{i-1}, \mathbf{h}^{\text{dec}}_{i-1}, \mathbf{c}_{i}),
\end{equation}
where Decoder$(\cdot)$ denotes the decoder network and $\mathbf{y}_{i-1}$ is the embedding of the preceding wordpiece. The model output can be calculated using a Softmax layer taking $\mathbf{h}^{\text{dec}}_i$ as input.

To calculate the TCPGen distribution in AED, the query combines the context vector $\mathbf{c}_i$ and the previously decoded token embedding $\mathbf{y}_{i-1}$ as shown in Eqn. \eqref{eq:TCPGeninlas_query}. 
\begin{equation}
    \mathbf{q}_i = \mathbf{W}^{\text{Q}}_c\mathbf{c}_i + \mathbf{W}^{\text{Q}}_y\mathbf{y}_{i-1},
    \label{eq:TCPGeninlas_query}
\end{equation}
where $\mathbf{W}^{\text{Q}}_c$ and $\mathbf{W}^{\text{Q}}_y$ are parameter matrices. The keys and values are computed from the decoder wordpiece embedding matrix as shown in Eqn. \eqref{eq:TCPGeninlas_key}.
\begin{equation}
    \mathbf{k}_j = \mathbf{W}^{\text{K}}\mathbf{y}_j \hspace{1cm} \mathbf{v}_j = \mathbf{W}^{\text{V}}\mathbf{y}_j,
    \label{eq:TCPGeninlas_key}
\end{equation}
where $\mathbf{y}_j$ denotes the $j$-th row of the embedding matrix. $\mathbf{W}^{\text{K}}$ and $\mathbf{W}^{\text{V}}$ are key and value parameter matrices which are shared throughout this paper. The TCPGen distribution and the TCPGen output can be computed using Eqns. \eqref{eq:TCPGen_attention} and \eqref{eq:TCPGen_value} respectively. The generation probability are calculated from the decoder hidden state $\mathbf{h}^{\text{dec}}_i$ and the TCPGen output $\mathbf{h}^{\text{ptr}}_i$, as shown in Eqn. \eqref{eq:las_gen}.
\begin{equation}
    {P}^{\text{gen}}_i = \sigma(\mathbf{W}^{\text{gen}}[\mathbf{h}^{\text{dec}}_i;\mathbf{h}^{\text{ptr}}_i]),
    \label{eq:las_gen}
\end{equation}
where $\mathbf{W}^\text{gen}$ is a parameter matrix. The distribution of $y_i$ can be calculated using Eqn. (\ref{eq:TCPGen_final}). In AED, deep biasing can be applied as shown in Eqn. (\ref{eq:deepbiasing_las}).
\begin{equation}
    P^{\text{mdl}}(y_i) = \text{Softmax}(\mathbf{W}^{\text{O}} [\mathbf{h}^{\text{dec}}_i; \mathbf{c}_i] + \mathbf{W}^{\text{db}} \mathbf{h}^{\text{db}}_i),
    \label{eq:deepbiasing_las}
\end{equation}
where the biasing vector $\mathbf{h}^{\text{db}}_i$ is obtained from the sum of embeddings of all wordpieces in $\mathcal{Y}^{\text{tree}}_i$, similar to \cite{deepshallow}.

\subsection{TCPGen in RNN-T}

A standard RNN-T consists of an encoder, a predictor and a joint network. The encoder in RNN-T is similar to that in AED which outputs $\mathbf{h}^{\text{enc}}_{1:T}$. The predictor encodes all wordpieces in the history into a vector, $\mathbf{h}^{\text{pred}}_i$, analogous to an LM. Given $K$ encoder outputs and $N$ predictor outputs, the joint network determines the output distribution $P(z_{i,t} | y_{1:i}, \mathbf{h}^{\text{enc}}_{t})$ for each combination of $i$ and $t$ as shown in Eqns. \eqref{eq:joint} and \eqref{eq:joint2}.
\begin{equation}
    \mathbf{h}^{\text{joint}}_{i, t} = \tanh(\mathbf{W}^{\text{joint}} [\mathbf{h}^{\text{pred}}_i; \mathbf{h}^{\text{enc}}_{t}]),
    \label{eq:joint}
\end{equation}
\begin{equation}
    P(z_{i,t} | y_{1:i-1}, \mathbf{h}^{\text{enc}}_{t}) = \text{Softmax}(\mathbf{W}^{\text{joint}}_2\mathbf{h}^{\text{joint}}_{i, t}),
    \label{eq:joint2}
\end{equation}
where the $\mathbf{W}^{\text{joint}}$ matrices are the parameter matrices of the joint network, and $z_{i,t} \in \mathcal{Y} \cup \{\varnothing\}$ where $\mathcal{Y}$ represents the set of all wordpieces. 

In RNN-T, the query for the TCPGen distribution is computed for each combination of $i, t$ as shown in Eqn. \eqref{eq:TCPGeninrnnt_query}.
\begin{equation}
    \mathbf{q}_{i,t} = \mathbf{W}^{\text{Q}}_c\mathbf{h}^{\text{enc}}_t + \mathbf{W}^{\text{Q}}_y\mathbf{y}_{i-1},
    \label{eq:TCPGeninrnnt_query}
\end{equation}
where $\mathbf{y}_{i-1}$ is the wordpiece embedding from the predictor. Key and value vectors are derived from the predictor embedding matrix. The generation probability for each $i, t$-pair is computed using the penultimate layer output of the joint network and the TCPGen output vector. 
\begin{equation}
    {P}^{\text{gen}}_{i,t} = \sigma(\mathbf{W}^\text{gen}[\mathbf{h}^{\text{joint}}_{i,t};\mathbf{h}^{\text{ptr}}_{i,t}])
\end{equation}
As $\varnothing$ only exists in $P^{\text{mdl}}$, its value is directly copied to the final output distribution as shown in Eqn. (\ref{eq:TCPGen_rnnt_interp})
\begin{equation}
P(z_{i,t}) = 
\begin{cases}
    P^{\text{mdl}}(\varnothing),& \text{if } z_{i,t} = \varnothing\\
    P(z_{i,t}),              & \text{otherwise}
\end{cases},
\label{eq:TCPGen_rnnt_interp}
\end{equation}
where $P(z_{i,t})$ is the interpolated probability for the output token $z_{i,t}$ in Eqn. (\ref{eq:TCPGen_final}), except that $P^{\text{ptr}}(z_{i,t})$ is scaled by a factor of $1-P^{\text{mdl}}(z_{i,t} \neq \varnothing)$ to ensure all probabilities sum up to 1. Moreover, whenever TCPGen is used in RNN-T, the biasing vector, $\mathbf{h}^{\text{ptr}}_{i,t}$ is always sent to the input of the joint network which yielded the best results as discussed in \cite{TCPGen}. As for AED, DB can be applied as described in \cite{deepshallow}.

\section{MBWE training for TCPGen}
\label{sec:mbr}

The MWE loss in end-to-end ASR systems minimises the expected value of word errors across all possible output sequences of a given input. Denoting the output sequence $Y = \{y_1,...,y_L\}$ and input sequence $X=\mathbf{x}_{1:T}$ for convenience, the MWE loss can be written as Eqn. \eqref{mbr1}.
\begin{equation}
    \mathcal{L}^{\text{mwe}}=\sum\nolimits_{Y} P(Y|X) \mathcal{W}^{\text{mwe}}(Y, {Y^*}),
    \label{mbr1}
\end{equation}
where ${Y^*}$ is the ground-truth output sequence and $\mathcal{W}^{\text{mwe}}(\cdot)$ is the risk function representing the number of word errors calculated using an edit-distance between each possible sequence $Y$ and the ground-truth sequence ${Y^*}$. The sum is performed over all possible sequences and $P(Y|X)$ is the probability of a specific sequence calculated from the end-to-end ASR model output. As it is intractable to enumerate over all possible sequences and calculate their probabilities, a common practice which has been widely adopted in MWE training for end-to-end ASR systems \cite{mwe0, mwe1, mwe2, mwe3, mwe4, mwe5} is to use the N-best hypotheses to approximate the expected word errors, as shown in Eqn. \eqref{mbr2}.
\begin{align}
    \mathcal{L}^{\text{mwe}} &\approx\frac{\sum_{Y_i\in \text{Beam}_{N}(X)}P(Y_i|X)\mathcal{W}^{\text{mwe}}(Y_i, {{Y}^*})}{\sum_{Y_i\in \text{Beam}_{N}(X)} P(Y_i|X)} \nonumber\\
    &=\sum\nolimits_{{Y}_i\in \text{Beam}_{N}({X})} \hat{P}({Y}_i|{X}) \mathcal{W}^{\text{mwe}}({Y}_i, {{Y}^*}),
    \label{mbr2}
\end{align}
where $\text{Beam}_{N}({Y})=\{{Y}_1,...,{Y}_N\}$ is the N-best hypotheses obtained via beam search, and $\mathcal{W}^{\text{mwe}}({Y}_i, {{Y}^*})$ represents the edit-distance function. The probabilities of the N-best hypotheses are normalised to form a valid distribution, where each normalised probability is represented as $\hat{P}({Y}_i|{X})$. Moreover, the average number of word errors over the N-best hypotheses is often subtracted from the number of word errors in each hypothesis as a form of variance reduction. 

To apply the MWE loss to contextual ASR with TCPGen which focuses on the correct recognition of biasing words, we propose a new risk function that includes an additional biasing word error term to the word error term as shown in Eqn. \eqref{mbr3}.
\begin{equation}
     \mathcal{W}^{\text{mbwe}}({Y}, {{Y}^*}) = \mu_1\mathcal{W}^{\text{mwe}}({Y}_i, {Y}^*) + \mu_2\mathcal{W}^{\text{bias}}({Y}_i, {Y}^*),
    \label{mbr3}
\end{equation}
where $\mathcal{W}^{\text{bias}}$ is the additional biasing word error term which is the edit-distance between the sequence of biasing words in ${Y}$ and the sequence of biasing words in ${Y}^*$. Scaling factors $\mu_1$ and $\mu_2$ control the importance of the word error term and the new biasing word error term. As a result, if $\mu_1=1$ and $\mu_2=0$, $\mathcal{W}^{\text{mbwe}}$ becomes $\mathcal{W}^{\text{mwe}}$. If $\mu_1=1$ and $\mu_2=1$, it is equivalent to giving a weight of 
2 to any rare word errors in the original word error rate. A new MBWE loss function is proposed to use  $\mathcal{W}^{\text{mbwe}}$ instead of  $\mathcal{W}^{\text{mwe}}$. That is, 
\begin{align}
    \nonumber\mathcal{L}^{\text{mbwe}}&=\sum_{Y} P(Y|X) \mathcal{W}^{\text{mbwe}}(Y, {Y^*})\\
    &\approx\sum\nolimits_{{Y}_i\in \text{Beam}_{N}({X})} \hat{P}({Y}_i|{X}) \mathcal{W}^{\text{mbwe}}({Y}_i, {{Y}^*}).
    \label{mbr4}
\end{align}
As a generic loss for end-to-end ASR systems, MBWE can be applied to the standard AED and RNN-T models, as well as other deep context models. 

\subsection{MBWE training in AED}

The MBWE loss can be applied to the AED model following a similar MWE training scheme proposed in \cite{mwe1} which also interpolated the MBWE loss with cross-entropy (CE) loss for better training stability, as shown in Eqn. \eqref{mbr_aed}.
\begin{equation}
    \mathcal{L} = \mathcal{L}^\text{mbwe} + \mathcal{L}^{\text{ce}},
    \label{mbr_aed}
\end{equation}
where $\mathcal{L}$ is the total loss function to be minimised, $\mathcal{L}^{\text{mbwe}}$ is the proposed MBWE loss in Eqn. \eqref{mbr4} and $\mathcal{L}^{\text{ce}}$ is the CE loss. As it is hard to train a randomly initialised model with the MBWE loss, which is similar to MWE, the MBWE loss is applied from the epoch when the model is reasonably trained with CE loss, which depends on optimisation algorithms. Moreover, to boost the efficiency of beam search which is the bottleneck in the time taken for training, batched beam search is implemented by parallelising the model forward computation of all beams of all utterances in the same mini-batch on a GPU.

\subsection{MBWE training in RNN-T}

The MBWE loss can also be applied to the RNN-T model following a similar MWE training scheme proposed in \cite{mwe2}, except that the original RNN-T loss is also included for stable training, as shown in Eqn. \eqref{mbr_rnnt}.
\begin{equation}
    \mathcal{L} = \mathcal{L}^\text{mbwe} + \mathcal{L}^{\text{rnn-t}},
    \label{mbr_rnnt}
\end{equation}
where $\mathcal{L}^{\text{rnn-t}}$ represents the original RNN-T loss \cite{e2e_rnnt_1}. Although previous work \cite{mwe2,mwe3} tried to obtain N-best lists using standard beam search for RNN-T, it is infeasible to perform such a training scheme on a single GPU with a limited number of CPUs, even with batched decoding. The major obstacle that restricts the level of parallel computation is the unknown number of wordpiece tokens to output at a given encoder step, as beams requiring two or more output tokens have to be handled separately. However, as the encoder output sequence is usually longer than the number of output wordpiece tokens, cases where two or more tokens are output at a specific encoder step, should be rare. To verify this conjecture, taking a standard RNN-T model trained on the Librispeech training set and decoded on its dev set as an example, the path taken and the number of output tokens at each encoder step for each 1-best hypothesis were recorded, as shown in Table \ref{tab:stats}.

\begin{table}[h]
    \centering
    \caption{Statistics of the Number of Output Wordpiece Tokens at Each Encoder Step for RNN-T 1-best Hypothesis on Librispeech Dev Set. Percentage is of the Total Number of Encoder Steps.}
    \begin{tabular}{ccccc}
    \toprule
       \# Wordpiece tokens  & 0 & 1 & 2+ \\
       \midrule
        Count & 80\% & 18\% & 2\%\\
    \bottomrule
    \end{tabular}
    \label{tab:stats}
\end{table}

As shown in the table, the vast majority of encoder steps output 0 or 1 wordpiece token, where 0 means a $\varnothing$ token is output. Although in the standard beam search, it is always required to compare the score with a second output token, it is often unnecessary for generating reasonably good N-best hypotheses, especially for MBWE training where a strong approximation using the N-best list has already been made. Therefore, a restricted beam search which only allows 0 or 1 output token at each encoder step is proposed for efficient MBWE training in RNN-T, which is similar to the one-step constrained beam search algorithm in \cite{nsc1} but with all neural network computations parallelised across all samples in the mini-batch. With restricted beam search, the model forward computation can be efficiently parallelised for all beams of all utterances in the same mini-batch on a single GPU.

\section{BLMD for TCPGen}
\label{sec:ilmd}

To incorporate an external LM, SF is often performed for both AED and RNN-T models via log-linear interpolation. Define the source domain data as the text of the training data for the end-to-end model, and the target domain data as the data used to train an external LM such that it generates better probability estimates for the test data. Then, the recognised sequence for LM SF can be written as Eqn. \eqref{eq:shallowfusion}. 
\begin{equation}
    Y^* = \underset{Y}{\text{arg max}}\,\log P^{\text{mdl}}(Y|X) + \alpha\,\log P^\text{tgt}(Y),
    \label{eq:shallowfusion}
\end{equation}
where $P^\text{mdl}(Y)$ is the output of the end-to-end system and $P^\text{tgt}(Y)$ is the probability of the output sequence predicted by an LM trained on the target domain. The interpolation factor $\alpha$ is a hyper-parameter to be determined. After decomposing the probability of each possible sequence $Y$ into a token-level sequence, $P^{\text{sf}}(y_i)$, the probability of each output token $y_i$ after SF, can be written as
\begin{equation}
    P^{\text{sf}}(y_i) = P^\text{mdl}(y_i){P^\text{tgt}(y_i)^{\alpha}},
    \label{eq:shallowfusion2}
\end{equation}
where the conditions on acoustic and history information were omitted for clarity. The density ratio method provides a Bayes' rule-grounded way to reduce the effect of the internal LM in the end-to-end system especially when there is a difference between the source and target domain data. That is,
\begin{align}
    P^{\text{sf}}(y_i) & = P^\text{mdl}(y_i)\frac{P^\text{tgt}(y_i)^{\alpha}}{P^\text{src}(y_i)^{\beta}},
    \label{eq:lmdisc}
\end{align}
where $P^\text{src}(Y)$ refers to the probability of the output sequence predicted by an LM trained on the source domain. The factors $\alpha$ and $\beta$ are hyper-parameters. In this way, the probabilities of commonly seen text patterns in the source domain are penalised, whereas those of text patterns specific to the target domain are boosted. Therefore, the density ratio LM discounting method can also be applied to the TCPGen component to further improve its performance on the biasing words that are rare in the source domain. As the final distribution comes from both the model and the TCPGen distribution which use different parameters and history information, density ratio SF is separately performed as 
\begin{align}
    P^{\text{sf}}(y_i) & = (1-P^\text{gen})P^\text{mdl}(y_i)\frac{P^\text{tgt}(y_i)^{\alpha_1}}{P^\text{src}(y_i)^{\beta_1}} \nonumber\\ & + P^\text{gen}P^\text{ptr}(y_i)\frac{P^\text{tgt}(y_i)^{\alpha_2}}{P^\text{src}(y_i)^{\beta_2}},
    \label{eq:lmdisc2}
\end{align}
where $P^\text{ptr}$ is the TCPGen distribution, and the same source and target LMs are used for both distributions, but with different sets of hyper-parameters $\alpha_1, \beta_1$ and $\alpha_2, \beta_2$. To avoid a complicated hyper-parameter search, the best set of $\alpha, \beta$ obtained from the standard end-to-end system can be directly applied to $\alpha_1, \beta_1$, and only $\alpha_2, \beta_2$ tuned to find the best values for the TCPGen distribution.

\section{Experimental setup}
\label{sec:exp}

\subsection{Data}

Experiments were performed on two different data sets, including the Librispeech audiobook corpus and the dialogue state-tracking challenge (DSTC) data. The Librispeech corpus \cite{librispeech} contains 960 hours of read English from audiobooks. 
The dev-clean and dev-other sets were held out for validation, and the test-clean and test-other sets were used for evaluation. Models trained on the Librispeech data were finetuned and tested on the DSTC data. 

The DSTC data was included as a realistic application with a limited amount of audio training resources of TCPGen where the ontology was used to extract contextual knowledge. The DSTC data contains human-machine task-oriented dialogues where user-side input audio was used for recognition. The DSTC track2 train and dev sets were used as the training and validation sets, and the DSTC track3 test set was used for evaluation. 
The 80-dim FBANK features at a 10~ms frame rate concatenated with 3-dim pitch features were used as the model input. SpecAugment \cite{specaug} with the setting $(W,F,m_F,T,p,m_T)=(40,27,2,40,1.0,2)$ was used without any other data augmentation or speaker adaptation.

\subsection{Biasing list selection}
For Librispeech, the full rare word list containing 200k distinct words proposed in \cite{DBRNNT} was used as the collection of all biasing words, in which more than 60\% were OOV words that did not appear in the Librispeech training set. Following the scheme in \cite{DBRNNT}, biasing lists were organised by finding words that belong to the full rare word list from the reference transcription of each utterance and adding a certain number of distractors to it. There were 10.3\% word tokens in the test sets that belong to the full rare word list and hence were covered by the biasing list during testing.

On the DSTC data, the biasing list arrangement by adding distractors was used for training only, where the full rare word list was augmented with words that occurred less than 100 times in the DSTC training data. During the evaluation, the ontology of DSTC3 which contains task-specific named entities was used to form the biasing list by extracting distinct words from those named entities and removing common words with 200 or more occurrences in the training data. This biasing list contained 243 distinct words, and 4.7\% word tokens in the test set belong to this biasing list. 


\vspace{-0.05cm}
\subsection{Model specification}
\vspace{-0.05cm}
Systems were built using the ESPnet toolkit \cite{espnet}. A unigram wordpiece model with 600 distinct wordpieces was built on the Librispeech data and was directly applied to the DSTC data. 
For both the AED and RNN-T models, a Conformer \cite{conformer} encoder was used. 
The Conformer encoder contained 16 conformer blocks consisting of 4 512-d attention heads. The AED additionally contained a single-layer 1024-d LSTM decoder and a 4-head 1024-d location-sensitive attention. The RNN-T additionally had a 1024-d predictor with a 1024-d single fully-connected layer joint network. 

For BLMD, a 2-layer 2048-d LSTM-LM trained on the Librispeech 800 million-word text training corpus was used as the target domain LM for Librispeech experiments. Each source domain LM trained on the text of the audio training data used a single-layer 1024-d LSTM. Each LM had the same wordpieces as the corresponding ASR system.

\vspace{-0.05cm}
\subsection{Training specifications}
\vspace{-0.05cm}
During training, biasing lists with 
1000 distractors were used for the Librispeech experiments and 100 distractors for the DSTC data. These lists were organised by finding biasing words from the reference and adding distractors. The dropping technique described in \cite{DBRNNT} was applied for AED model training, where biasing words that were found in the reference transcription had a 30\% probability to be removed from the biasing list. This was to prevent the model from being over-confident about TCPGen outputs. 
The Noam \cite{transformer} optimiser was used for the Conformer. 
The MBWE loss was applied after 16 epochs. The beam size for MBWE training was 5 for all experiments and 30 for decoding. A coverage penalty \cite{coverage_p} of 0.01 was applied to AED models.

\vspace{-0.05cm}
\subsection{Evaluation metrics}
\vspace{-0.05cm}
In addition to WER, a rare word error rate (R-WER) was used to evaluate the system performance on biasing words that were ``rare" in the training data for that system. R-WER is the total number of \textit{error} word tokens that belong to the biasing list divided by the total number of word tokens in the test set that belong to the biasing list. Insertion errors were counted in R-WER if the inserted word belonged to the biasing list \cite{DBRNNT}. As WERs on Librispeech test sets were all small, for the rest of this paper, 2 decimal places were included for WER whereas one decimal place was used for other results. Moreover, for small WER and R-WER reductions, a project-by-project sign test was performed for Lirbispeech experiments based on the ``project ID" of each utterance. The same sign test was performed dialogue-by-dialogue for the DSTC data.

\section{Results}
\label{sec:results}

\begin{figure*}[t]
    \centering
    \includegraphics[scale=0.27]{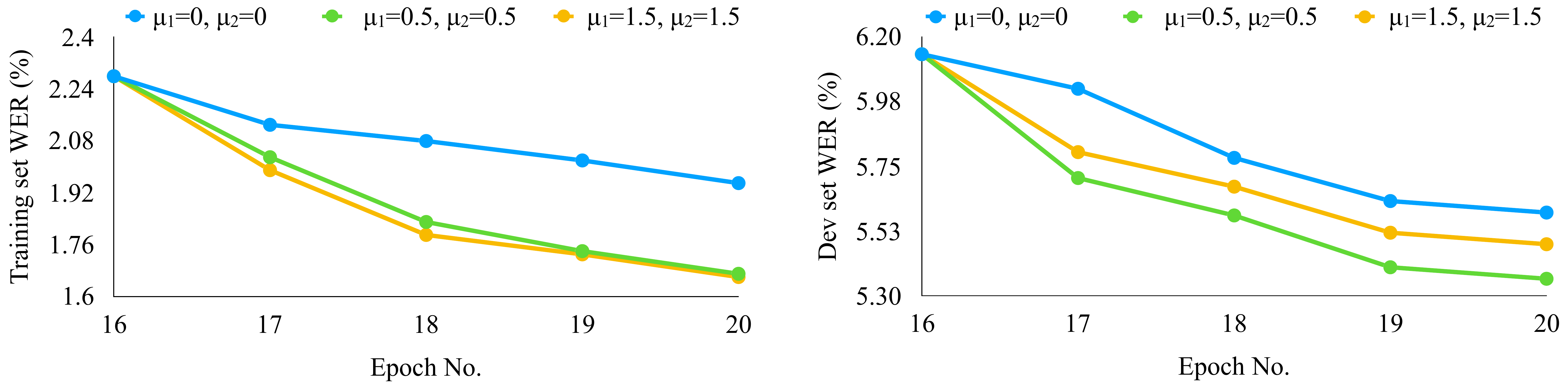}
    \caption{Plots of training (left) and dev (right) set WERs across 4 training epochs. Training set WER was calculated on 5\% randomly sampled utterances from the full 960-hour training set. Dev-set combines both dev-clean and dev-other sets. MBWE parameters $\mu_1, \mu_2$ were defined in Eqn. \eqref{mbr3}.}
    \label{fig:bmbrtrain}
\end{figure*}

\subsection{Conformer AED experiments on Librispeech}
The TCPGen component together with proposed MBWE and BLMD algorithms was first applied to the conformer AED model. 
\begin{table*}[t]
    \centering
    \caption{WER and R-WER on Librispeech test-clean and test-other sets for Conformer-based systems trained on Librispeech full 960-hour training set. MBWE params. include $\mu_1$ and $\mu_2$ in Eqn. \eqref{mbr3}, and BLMD params include $\alpha_1,\beta_1$ and $\alpha_2, \beta_2$ in Eqn. \eqref{eq:lmdisc2}. The baseline here refers to the standard Conformer AED model.}
    \begin{tabular}{lcccccc}
    \toprule
    & & & \multicolumn{2}{c}{test-clean} & \multicolumn{2}{c}{test-other} \\
    System & MBWE params. & BLMD params. & \%WER & \%R-WER & \%WER & \%R-WER \\
    \midrule
    Baseline & $\mu_1=0.0, \mu_2=0.0$ & - & 3.71 & 13.2 & 9.36 & 29.5 \\
    Baseline & $\mu_1=0.5, \mu_2=0.0$ & - & 3.65 & 13.0 & 9.02 & 28.9\\
    Baseline & $\mu_1=0.5, \mu_2=0.5$ & - & {3.62} & {12.8} & 9.08 & {28.6} \\
    \midrule
    TCPGen & $\mu_1=0.0, \mu_2=0.0$ & - & 3.23 & 8.6 & 8.43 & 21.3 \\
    TCPGen & $\mu_1=0.5, \mu_2=0.5$ & - & \textbf{2.96} & \textbf{7.6} & \textbf{7.88} & \textbf{19.5} \\
    \midrule
    \midrule
    Baseline & $\mu_1=0.0, \mu_2=0.0$ & $\alpha_1=0.5, \beta_1=0.3$ & 3.33 & 12.3 & 8.04 & 27.6 \\
    Baseline & $\mu_1=0.5, \mu_2=0.0$ & $\alpha_1=0.5, \beta_1=0.3$ & 3.19 & 12.2 & 7.95 & 27.1 \\
    Baseline & $\mu_1=0.5, \mu_2=0.5$ & $\alpha_1=0.5, \beta_1=0.3$ & 3.17 & 11.7 & 7.92 & 27.3 \\
    \midrule
    TCPGen & $\mu_1=0.0, \mu_2=0.0$ & $\alpha_1=0.5, \beta_1=0.3, \alpha_2=0.3, \beta_2=0.3$ & 2.79 & 6.9 & 7.40 & 19.5 \\
    TCPGen & $\mu_1=0.5, \mu_2=0.5$ & $\alpha_1=0.5, \beta_1=0.3, \alpha_2=0.3, \beta_2=0.3$ & \textbf{2.59} & \textbf{6.4} & \textbf{7.13} & \textbf{18.2} \\
    \bottomrule
    \end{tabular}
    \label{tab:aedcfm}
\end{table*}
With the Noam optimiser, the learning rate was a smooth function and preliminary experiments found that adjusting the weight of the cross-entropy loss yielded significantly worse results as it effectively introduced an abrupt change to the learning rate. Therefore, to adjust the contribution of the MBWE loss, different values of $\mu_1$ and $\mu_2$ were used while keeping the coefficient of the cross-entropy loss to 1. The effect of using small and large values of $\mu_1$ and $\mu_2$ on the training and dev set WER are shown in Fig. \ref{fig:bmbrtrain}. As shown in Fig. \ref{fig:bmbrtrain}, applying MBWE with both small and large values had a similar effect on the training set WER, whereas using smaller values produced better results on the dev set and hence was adopted for the experiments.

\begin{figure}[t]
    \centering
    \includegraphics[scale=0.27]{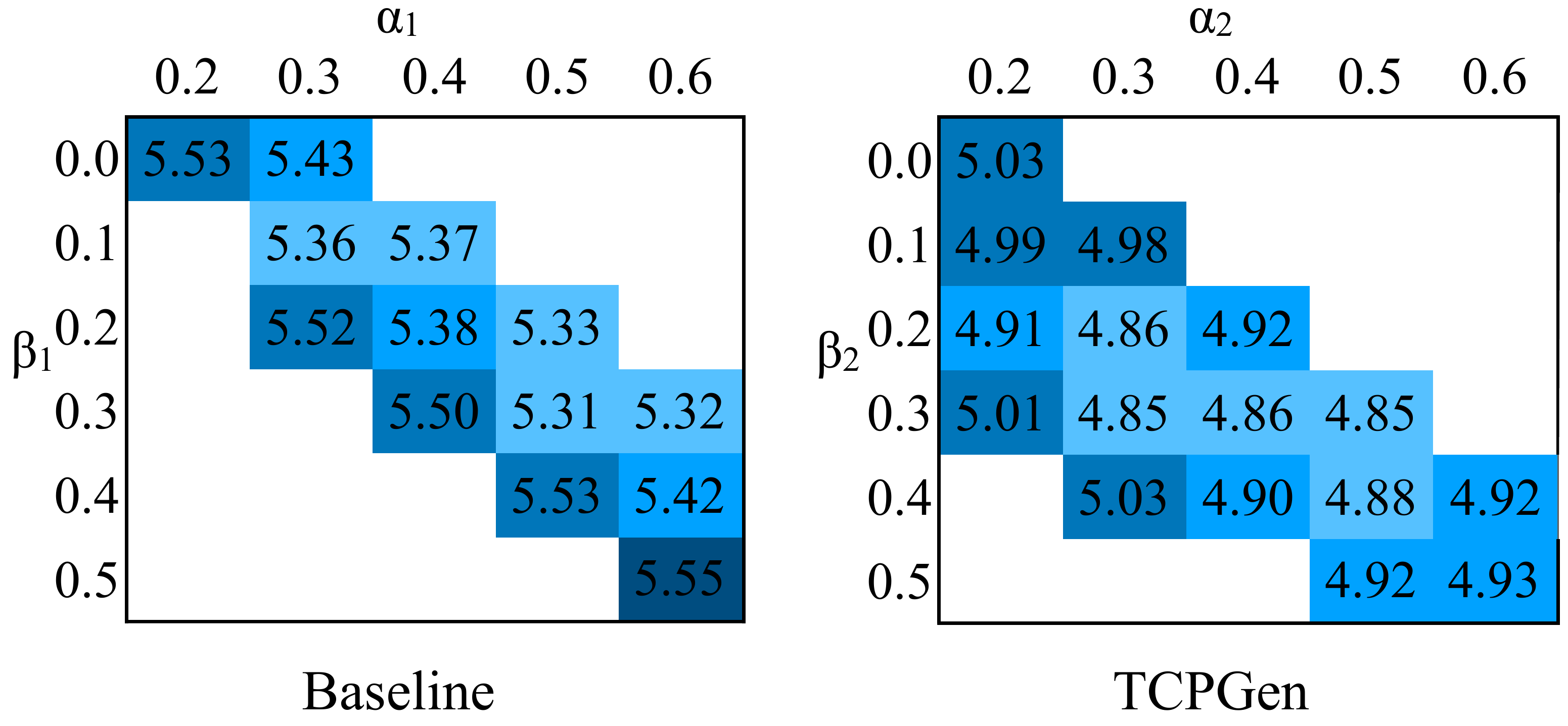}
    \caption{Illustration of tuning BLMD hyper-parameters for the baseline standard Conformer AED model and the Conformer AED model with TCPGen. Numbers in each grid are dev set WER in percentage. Left: Tuning $\alpha_1, \beta_1$ on the baseline model. Right: Tuning $\alpha_2, \beta_2$ on the TCPGen model with the best set of $\alpha_1, \beta_1$ found from the baseline on the left.}
    \label{fig:blmdtune}
\end{figure}

Then, the best set of BLMD parameters was searched for and applied to the trained models during decoding. The search procedure is illustrated in Fig. \ref{fig:blmdtune}. The left part of Fig. \ref{fig:blmdtune} shows the dev set WER of different sets of BLMD parameters $\alpha_1, \beta_1$ for the baseline standard AED system. The best values found here were $\alpha_1=0.5, \beta=0.3$, which were then fixed for the search of $\alpha_2, \beta_2$, as shown on the right part of Fig. \ref{fig:blmdtune} for the system with TCPGen. As a result, $\alpha_2=0.3, \beta_2=0.3$ were used which indicates that a stronger LM discounting effect was needed for the TCPGen distribution. 

The results for the Conformer AED model are summarised in Table \ref{tab:aedcfm}. As shown in Table \ref{tab:aedcfm}, for the baseline standard Conformer AED system, using the MBWE loss benefits the R-WER. When MBWE is applied to TCPGen, there was a 12\% relative reduction in R-WER on the test-clean set and 9\% on the test-other set compared to the TCPGen system without MBWE training. This increased the relative R-WER improvement by using TCPGen from 33\% to 41\% on the test-clean set, and from 28\% to 32\% on the test-other set compared to the baseline with the same training loss (i.e. comparing row 1 with row 4 and row 3 with row 5 in Table \ref{tab:aedcfm}).

After applying BLMD, obvious reductions in R-WER were observed for both the baseline and TCPGen systems on both test sets. In particular, large R-WER reductions were found when different discounting factors were applied to the TCPGen distribution, which further increased the relative R-WER improvement by using TCPGen from 41\% to 46\% on the test-clean set and from 32\% to 33\% on the test-other set (comparing row 3 with row 5 and row 8 with row 10).

\begin{table*}[t]
    \centering
    \caption{WER and R-WER on Librispeech test-clean and test-other sets for Conformer-based RNN-T models trained on Librispeech full 960-hour training set. MBWE params. include $\mu_1$ and $\mu_2$ in Eqn. \eqref{mbr3}, and BLMD params include $\alpha_1,\beta_1$ and $\alpha_2, \beta_2$ in Eqn. \eqref{eq:lmdisc2}. The baseline here refers to the standard Conformer RNN-T model.}
    \begin{tabular}{lcccccc}
    \toprule
    & & & \multicolumn{2}{c}{test-clean} & \multicolumn{2}{c}{test-other} \\
    System & MBWE params. & BLMD params. & \%WER & \%R-WER & \%WER & \%R-WER \\
    \midrule
    Baseline & $\mu_1=0.0, \mu_2=0.0$ & - & 4.02 & 14.1 & 10.12 & 33.1\\
    Baseline & $\mu_1=0.5, \mu_2=0.0$ & - & 4.01 & 14.0 & 9.96 & 32.5 \\
    Baseline & $\mu_1=0.5, \mu_2=0.5$ & - & 3.87 & 13.8 & 9.80 & 31.8 \\
    \midrule
    DB & $\mu_1=0.0, \mu_2=0.0$ & - & 3.57 & 10.4 & 9.45 & 25.0 \\
    TCPGen & $\mu_1=0.0, \mu_2=0.0$ & - & 3.40 & 8.9 & 8.79 & 22.2 \\
    TCPGen & $\mu_1=0.5, \mu_2=1.0$ & - & \textbf{3.12} & \textbf{8.1} & \textbf{8.64} & \textbf{20.7} \\
    \midrule
    \midrule
    Baseline & $\mu_1=0.0, \mu_2=0.0$ & $\alpha_1=0.4, \beta_1=0.1$ & 3.55 & 12.5 & 8.90 & 30.4 \\
    Baseline & $\mu_1=0.5, \mu_2=0.0$ & $\alpha_1=0.4, \beta_1=0.1$ & 3.38 & 12.6 & 8.59 & 29.5 \\
    Baseline & $\mu_1=0.5, \mu_2=0.5$ & $\alpha_1=0.4, \beta_1=0.1$ & 3.28 & 12.2 & 8.50 & 28.8 \\
    \midrule
    TCPGen & $\mu_1=0.0, \mu_2=0.0$ & $\alpha_1=0.4, \beta_1=0.1, \alpha_2=0.2, \beta_2=0.1$ & 3.02 & 8.0 & 7.49 & 18.6 \\
    TCPGen & $\mu_1=0.5, \mu_2=1.0$ & $\alpha_1=0.4, \beta_1=0.1, \alpha_2=0.1, \beta_2=0.0$ & \textbf{2.79} & \textbf{7.0} & \textbf{7.44} & \textbf{18.2} \\
    \bottomrule
    \end{tabular}
    \label{tab:rnntcfm}
\end{table*}

\subsection{Conformer RNN-T experiments on Librispeech}
Experiments were then performed on Librispeech full 960-hour training data as shown in Table \ref{tab:rnntcfm}. Preliminary experiments on Librispeech found that $\mu_2$ for the MBWE loss should be set larger than $\mu_1$ for better performance when using TCPGen in RNN-T. The baseline here is a standard Conformer-based RNN-T model. The MBWE and BLMD hyper-parameters were found in the same way as for the AED experiments. In addition to the standard baseline system, the DB method proposed in \cite{DBRNNT} was used as a biasing method for comparison. In general, consistent and significant WER and R-WER reductions were achieved using TCPGen compared to both the baseline and the DB system, with a p-value less than $.001$. MBWE with a restricted beam search achieved WER and R-WER reductions for both baseline and TCPGen systems. In particular, the relative R-WER improvement increased from 37\% to 41\% on the test-clean set and from 33\% to 37\% on the test-other set compared to the baseline system (comparing row 4 to row 1 and row 6 to row 3 in Table \ref{tab:rnntcfm}). 

Applying BLMD achieved further WER and R-WER reductions for all systems. In particular, BLMD increased the relative R-WER reduction from 41\% to 43\% on test-clean and from 37\% to 38\% on test-other (comparing row 6 to row 3 and row 11 to row 9 in Table \ref{tab:rnntcfm}). The sign test was used to compare TCPGen with and without MBWE training, as the WER and R-WER were smaller than those observed in AED. All R-WER improvements after applying MBWE, including those when BLMD was applied, were significant at $p<.05$.

\begin{figure}
    \centering
    \includegraphics[scale=0.3]{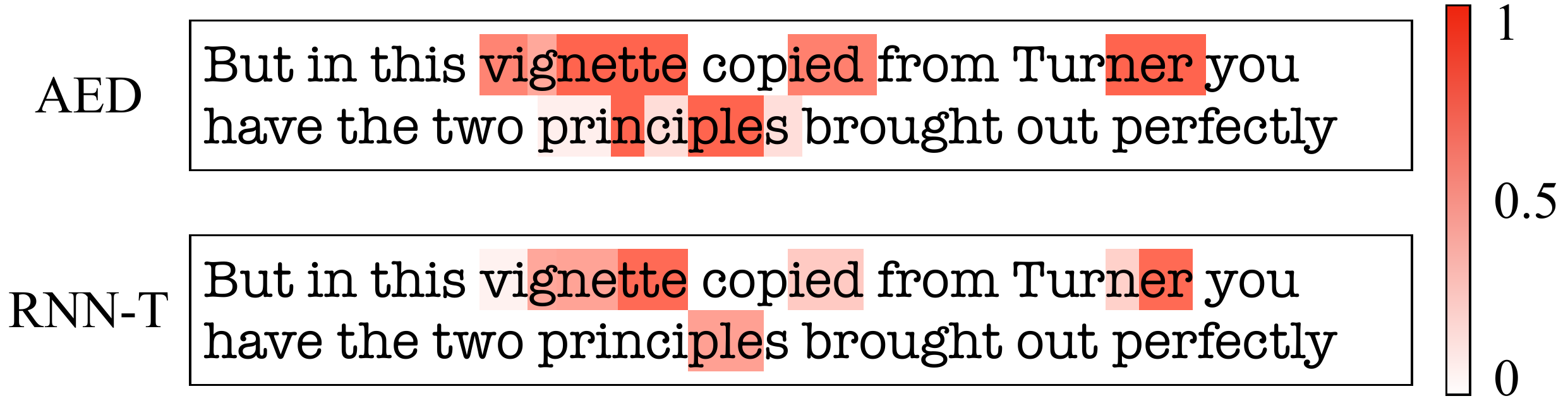}
    \caption{Heat map of the generation probability for each wordpiece in an utterance taken from recognition results to show how each system spots where to use contextual biasing. Biasing words are \texttt{vignette} and \texttt{Turner}.}
    \label{fig:compare}
\end{figure}

Compared to the results for the AED model, the R-WER improvement was smaller in general, and to investigate this discrepancy, the generation probabilities $P^\text{gen}$ for TCPGen in AED and RNN-T models are plotted in Fig. \ref{fig:compare} for an example utterance from the test-clean set.

As shown in Fig. \ref{fig:compare}, the colour on biasing words was lighter for RNN-T than AED, indicating that RNN-T had a smaller dependency on TCPGen than AED. This resulted in a smaller R-WER reduction using TCPGen in RNN-T. During RNN-T training, the probability of each alignment to be maximised contained only a tiny portion of the factorised probabilities that correspond to a new wordpiece output with a large portion of the blank symbol, $\varnothing$, output, and TCPGen has an effect on only a small portion of those that output a new wordpiece token. Therefore, RNN-T used TCPGen much less frequently than AED during training. Moreover, when TCPGen was used, the generation probability was scaled by $1-P(\varnothing)$ to ensure a valid output distribution. To encourage RNN-T to use TCPGen, the scaling factor for the rare word error loss for TCPGen was set to 5.0 during MBWE training.

\begin{table}[t]
    \centering
    \caption{Zero-shot WERs on OOV words on the combined test-clean and test-other set using the baseline and TCPGen systems with MBWE and BLMD. Same biasing lists were used as those in Table \ref{tab:aedcfm} and Table \ref{tab:rnntcfm}.}
    \begin{tabular}{lcc}
    \toprule
      Systems   &  AED & RNN-T \\
      \midrule
      Baseline + MBWE + BLMD  & 75.6\% & 78.0\% \\
      TCPGen + MBWE + BLMD   & \textbf{37.8}\% & \textbf{41.7}\% \\
      \bottomrule
    \end{tabular}
    \label{tab:oovwer}
\end{table}

TCPGen also achieved \textit{zero-shot learning} of OOV words incorporated in the biasing list. There were 468 OOV word tokens in the combined test-clean and test-other set that were covered by the biasing list. The OOV WER which was measured in the same way as R-WER but for OOV words on the combined test sets was separately reported in Table \ref{tab:oovwer} for the baseline and for TCPGen systems using MBWE and BLMD. Note that these systems used exactly the same biasing lists with 1000 distractors as those in Table \ref{tab:aedcfm} and Table \ref{tab:rnntcfm}, so they had the same WER and R-WER as those corresponding systems. As a result, TCPGen achieves a large OOV WER reduction compared to the best baseline system, and the majority of OOV words could be correctly recognised once incorporated in the biasing list.

\subsection{DSTC experiments}

Finally, TCPGen and the proposed MBWE and BLMD methods were evaluated on the DSTC data where the biasing list was extracted from the ontology. Models trained on the Librispeech 960-hour data were finetuned on the DSTC track2 training set. For the baseline and TCPGen systems without MBWE training, finetuning was performed only with the CE loss, whereas for systems trained with MBWE, the MBWE loss was also applied during finetuning. The hyper-parameters for MBWE were found in the same way as before, with $\mu_1=0.5$ and $\mu_2=5.0$. Moreover, as it is difficult to obtain large external task-oriented dialogue data for LM training, an LM was trained only on the DSTC2 training data to perform either shallow fusion or LM discounting. For the baseline system, this DSTC LM was found to be more effective as an SF LM with $\alpha_1=0.1, \beta_1=0.0$, which achieved a limited WER improvement. For TCPGen, BLMD was applied with the same $\alpha_1$ and $\beta_1$ as the baseline, and $\alpha_2=0.0, \beta_2=0.1$, which, in addition to the SF, the internal LM effect was discounted in the TCPGen distribution. The WER and R-WER are reported in Table \ref{tab:dstcexp} where the R-WER was measured for biasing words that appeared in the ontology. 

\begin{table}[h]
    \centering
    \caption{WER and R-WER (in brackets) on the DSTC3 test set for Conformer AED and RNN-T models trained on Librispeech full 960-hour training set and finetuned on DSTC2 train and dev sets.}
    \begin{tabular}{lccc}
    \toprule
    System &  AED(\%) & RNN-T(\%) \\
    \midrule
    Baseline & 21.31 (61.5) & 21.26 (64.2) \\
    Baseline + MBWE & 20.81 (60.4) & 21.15 (64.1) \\
    Baseline + MBWE + SF & 20.73 (60.4) & 20.63 (64.1) \\
    \midrule
    TCPGen & 20.38 (45.2) & 20.05 (49.2)\\
    TCPGen + MBWE & {20.00} (43.6) & 19.87 ({47.4})\\
    TCPGen + MBWE + BLMD & \textbf{19.74} (\textbf{40.3}) & \textbf{19.13} (\textbf{40.9})\\ 
    \bottomrule
    \end{tabular}
    \label{tab:dstcexp}
\end{table}

Progressive WER and R-WER reductions were achieved by applying MBWE and BLMD successively for both AED and RNN-T using TCPGen. For AED, using TCPGen achieved a 26\% relative R-WER reduction compared to the baseline, which then increased to 33\% with BLMD for AED. For RNN-T, TCPGen alone achieved a 23\% relative R-WER reduction compared to the baseline, which increased to 36\% relative R-WER reduction compared to the corresponding baseline with SF. Moreover, a dialogue-by-dialogue sign test was performed between TCPGen and TCPGen with MBWE loss. For both AED and RNN-T, R-WER improvements were significant at a p-value less than 0.05.


\section{Conclusion}
\label{sec:conclusion}
This paper proposed the TCPGen component for contextual ASR. TCPGen combines a neural pointer generator with a symbolic prefix-tree search. Meanwhile, the minimum biasing word error (MBWE) loss was proposed to improve the training of TCPGen with an emphasis on biasing words, and the biasing-word-driven LM discounting (BLMD) method was proposed for decoding to account for the internal LM effect. Experiments were performed on Librispeech and DSTC dialogue data.
Consistent and significant WER improvements were found using TCPGen, especially on biasing words. Applying MBWE and BLMD achieved further significant R-WER reductions compared to the original TCPGen reductions.


\vfill

\end{document}